\pdfoutput=1                                    
\documentclass[a4paper, 10pt, conference]{ieeeconf}      
\usepackage{FG2021}
\usepackage{graphicx}
\usepackage{multirow}
\usepackage{amsmath}
\usepackage{amssymb}
\usepackage{hyperref}
\usepackage{amsfonts}
\usepackage{mathfixs}
\hypersetup{
    colorlinks=true,
    linkcolor=blue,
    filecolor=magenta,      
    urlcolor=blue,
    pdftitle={Overleaf Example},
    pdfpagemode=FullScreen,
    }
\FGfinalcopy 

\IEEEoverridecommandlockouts                              
\def\FGPaperID{400} 

\title{\LARGE \bf
Cross Attentional Audio-Visual Fusion for Dimensional \\Emotion Recognition
}

\author{\parbox{16cm}{\centering
    {\large R. Gnana Praveen, Eric Granger and Patrick Cardinal}\\
    {\normalsize 
    Laboratoire d’imagerie, de vision et d’intelligence artificielle (LIVIA) 
    \\ École de technologie supérieure, Montreal, Canada
    }}
}

\begin{document}

%
%
%





\ifFGfinal
\thispagestyle{empty}
\pagestyle{empty}
\else
\author{Anonymous FG2021 submission\\ Paper ID \FGPaperID \\}
\pagestyle{plain}
\fi
\maketitle

\begin{abstract}
Multimodal analysis has recently drawn much interest in affective computing, since it can improve the overall accuracy of emotion recognition over isolated uni-modal approaches. The most effective techniques for multimodal emotion recognition efficiently leverage diverse and complimentary sources of information, such as facial, vocal, and physiological modalities, to provide comprehensive feature representations. In this paper, we focus on dimensional emotion recognition based on the fusion of facial and vocal modalities extracted from videos, where complex spatiotemporal relationships may be captured. Most of the existing fusion techniques rely on recurrent networks or conventional attention mechanisms that do not effectively leverage the complimentary nature of audio-visual (A-V) modalities. We introduce a cross-attentional fusion approach to extract the salient features across A-V modalities, allowing for accurate prediction of continuous values of valence and arousal.  Our new cross-attentional A-V fusion model efficiently leverages the inter-modal relationships. In particular, it computes cross-attention weights to focus on the more contributive features across individual modalities, and thereby combine contributive feature representations, which are then fed to fully connected layers for the prediction of valence and arousal. The effectiveness of the proposed approach is validated experimentally on videos from the RECOLA and Fatigue (private) data-sets. Results indicate that our cross-attentional A-V fusion model is a cost-effective approach that outperforms state-of-the-art fusion approaches. Code is available: \url{https://github.com/praveena2j/Cross-Attentional-AV-Fusion}
\end{abstract}

\section{INTRODUCTION}
Automatic recognition of emotions is an important task that facilitates natural interaction between human and machine. Emotion recognition is found in many applications, such as assessment of anger, fatigue, depression, pain, motivation, and stress in health care, e-learning, security, etc. Recognizing emotions is a challenging problem in real world scenarios as expressions linked to an emotional state are often diverse in nature across individuals and cultures \cite{Anagnostopoulos}. Human emotions can be conveyed through various modalities such as face, voice, text and physiology (electroencephalogram, electrocardiogram, etc.), which typically carry complementary information among them. Although human emotions can be expressed through various modalities, vocal and facial modalities are the predominant contact-free channels through which they can be efficiently expressed. Therefore, audio-visual (A-V) fusion for emotion recognition has been widely explored for several decades \cite{wu_lin_wei_2014}. Depending on the type of labels, emotion recognition can be formulated as a discrete classification problem (e.g., a person eliciting happy or sad emotions), or as a continuous regression problem (e.g. continuous values of valence and arousal). Though classification conveys the type of emotion being expressed, it fails to capture the wide range of emotions on a finer granularity. In this paper, we focus on dimensional A-V emotion recognition based on videos in the context of regression. Valence and arousal are widely used for estimating emotion intensities in continuous domain, where valence spans the wide range of emotions from sad to happy, and arousal reflects the energy or intensity of the emotions.

Typically, A-V fusion for emotion recognition can be achieved by three major strategies: decision-, feature-, and model-level fusion \cite{wu_lin_wei_2014}. In decision-level fusion (late fusion), multiple modalities are trained end-to-end independently, and then the predictions obtained from the individual modalities are fused to obtain the final predictions. Although decision-level fusion is easy to implement, and requires less training, it neglects the interactions across the individual modalities, thereby resulting in limited improvement over uni-modal approaches. Conventionally, feature-level fusion (early fusion) is achieved by concatenating the features of A-V modalities immediately after they are extracted, which is further used for predicting the final outputs. Though feature-level fusion allows interaction between the modalities at the low level features, it fails to leverage the interactions between the A-V features (inter-modal relationships) across the individual modalities, thereby resulting in limited improvement in performance \cite{wu_lin_wei_2014}. Model-level fusion is the most effective way to leverage the complimentary nature of the modalities to obtain comprehensive feature representation. The interactions between A-V features such as intra- or inter-modal relationships are explicitly captured, usually based on models, e.g., deep networks \cite{cite3}, hidden Markov models \cite{cite4}, and kernel methods \cite{cite5}. Inspired by the performance of model-based fusion and deep networks, we focus on deep multimodal representation learning for A-V dimensional emotion recognition to efficiently leverage the correlation across A and V modalities.

Deep learning (DL) models provide state-of-the-art performance in many V recognition applications, such as object detection, action recognition, etc. Inspired by their performance, several approaches have been proposed for video-based dimensional emotion recognition using CNNs to obtain the deep features, and a recurrent neural network to capture the temporal dynamics \cite{cite6,cite7}. In most of these approaches \cite{cite7, cite8}, A-V fusion is performed by concatenating the deep features extracted from individual facial and vocal modalities, and fed to LSTM for predicting valence and arousal. Although LSTM based fusion models the intra-modal relationship (temporal) and improves the performance of the system, it does not effectively capture the inter-modal relationships across the individual modalities. We therefore investigate the prospect of extracting more contributive features across A and V modalities in order to leverage their complimentary temporal relationships. 

Attention mechanisms have recently gained much interest in the areas of computer vision and machine learning as they allow extracting task relevant features, thereby improving system performance. This has been extensively explored for various applications, such as event/action recognition \cite{9157522}, emotion recognition \cite{Jiyoung}, etc. Most of the existing attention based approaches for dimensional emotion recognition explore the intra-modal relationships \cite{Jiyoung}. Although a few approaches \cite{srini_2021_SLT, cite8} attempt to capture the cross-modal relationships using cross-attention based on transformers, they fail to effectively leverage the complimentary relationship of A-V modalities. Indeed, their computation of attention weights does not consider the correlation among the A and V features. In this paper, we introduce a cross-attentional A-V fusion model for predicting valence and arousal by exploiting the interactions between A-V features. By estimating the cross-correlation across deep A and V features, the proposed model employs the cross-correlation matrix to capture the relevant complimentary information for accurate dimensional emotion recognition. The cross-correlation based attention helps to capture the semantic relevance between A and V features, thereby effectively leveraging the complimentary A-V relationships. Though cross-attention based on cross-correlation have been explored for other applications such as few shot classification \cite{NEURIPS2019_01894d6f} and action classification \cite{lee2021crossattentional}, we investigate it in the context of regression for dimensional emotion recognition.  

\noindent \textbf{Main contributions: }
(1) We propose a cross-attentional A-V fusion model based on cross-correlation to effectively exploit the complimentary relationship across modalities for dimensional emotion recognition.
(2) Unlike prior approaches, we leverage the interactions between A-V features (inter-modal relationships) to obtain complimentary representations for dimensional emotion recognition.
(3) For proof-of-concept, we consider Inflated 3D CNN model \cite{8099985} to efficiently extract the spatiotemporal features for the facial modality, coupled with a 2D-CNN model to extract A features from a spectrogram representation for the vocal modality. Experimental results on RECOLA and Fatigue (private) data-sets shows that our proposed cross-attentional A-V fusion can outperform state-of-the-art fusion models for dimensional emotion recognition.

\section{Related Work}
\noindent \textbf{A) A-V Fusion for Dimensional Emotion Recognition:}
One of the early approaches using DL models for A-V fusion based dimensional emotion recognition was proposed by Tzirakis et al. \cite{cite7}, where A and V features are obtained using ResNet50 and 1D CNN respectively. The obtained features are then concatenated and fed to Long short-term memory model (LSTM) for the prediction of valence and arousal. Juan et al. \cite{8914655} investigated an empirical study of fine-tuning pretrained CNN models by freezing various convolutional layers. Schonevald et al. \cite{cite6} explored knowledge distillation using student-teacher model for V modality and CNN model for A modality using spectrograms. The deep feature representations are combined using a model-based fusion strategy, where RNNs are used to capture the temporal dynamics. Inspired by the deep auto-encoders, Nguyen et al. \cite{9374787} investigated the prospect of how to simultaneously learn compact representative features from A and V modalities using deep auto-encoders. They have proposed a deep model of two-stream auto-encoders and LSTM for efficiently integrating V and A streams for dimensional emotion recognition. Though the above mentioned approaches have shown significant improvement for dimensional emotion recognition, they fail to capture the inter-modal relationships and relevant salient features specific to the task. Therefore, we have focused on capturing the comprehensive features in a complimentary fashion using attention mechanisms.  

\noindent \textbf{B) Attention for A-V Fusion:}
Attention mechanisms are widely used in the context of multimodal fusion with various modalities such as A and text \cite{Lee2020,N.2020}, V and text \cite{8578827, Wei_2020_CVPR}, etc. In this work, we focus on attention mechanisms for model-based A-V fusion in videos. Zhao et al. \cite{Zhao_Ma_Gu_Yang_Xing_Xu_Hu_Chai_Keutzer_2020} proposed an end-to-end architecture for emotion classification by integrating spatial, channel-wise and temporal attentions into V network and temporal attention into A network. Esam et al. \cite{9191019} explored attention to weigh the time windows of a video sequence to efficiently exploit the temporal interactions between the A-V modalities. They used transformer \cite{NIPS2017_3f5ee243} based encoders to obtain the attention weights through self attention for emotion classification. Lee et al. \cite{Jiyoung} proposed spatiotemporal attention for the V modality to focus on emotional salient parts using Convolutional LSTM (ConvLSTM) modules and a temporal attention network using deep networks for A modality. Then the attended features are concatenated and fed to the regression network for the prediction of valence and arousal. However, these approaches focus on modeling the intra-modal relationships and fail to exploit the complimentary nature of the A-V modalities.

Wang et al. \cite{Wang_ICMI_2020} investigated the prospect of exploiting the implicit contextual information along with the A and V modalities. They have proposed an end-to-end architecture using cross-attention based on transformers for A-V group emotion recognition. Srinivas et al. \cite{srini_2021_SLT} also explored transformers with cross modal attention for dimensional emotion recognition, where cross-attention is integrated along with self attention. Tzirakis et al. \cite{cite8} investigated various fusion strategies along with attention mechanisms for A-V fusion based dimensional emotion recognition. They have further explored self attention as well as cross-attention fusion based on transformers in order to enable the extracted features of different modalities to attend to each other. Although these approaches have explored cross-attention with transformers, they fail to leverage the cross-correlation based semantic relevance among the A-V features.

Wang et al. \cite{9197622} addressed the problem of multimodal feature fusion along with frame alignment issues between A and V modalities using cross-attention for speech recognition. Unlike prior approaches based on transformers, we use a simple yet efficient attention mechanism based on cross-correlation across A and V modalities. The cross-correlation across A and V features helps to effectively retain the semantic relevance and capture the complimentary relationships among the A-V modalities. Jun et al. \cite{lee2021crossattentional} is the most similar to our approach, where multi-stage cross-attentional A-V fusion is used to collaboratively fuse A and V features for localizing and classifying actions in videos. The proposed approach primarily differs from \cite{lee2021crossattentional} in two respects: (1) cross-attentional fusion is explored in an iterative mechanism in \cite{lee2021crossattentional}, whereas we apply cross-attention in a single iteration, and still improve system performance. (2) In \cite{lee2021crossattentional}, cross-attentional fusion is used for action recognition in the context of classification, whereas we adapt it to dimensional emotion recognition in the context of regression.

\section{Proposed Approach}
In this section, we introduce the cross-attentional A-V fusion model that extracts complimentary features across facial and vocal modalities, thereby providing a comprehensive representation to improve the overall performance. 
\begin{figure*}[t!]
\centering
\includegraphics[width=1.0\linewidth]{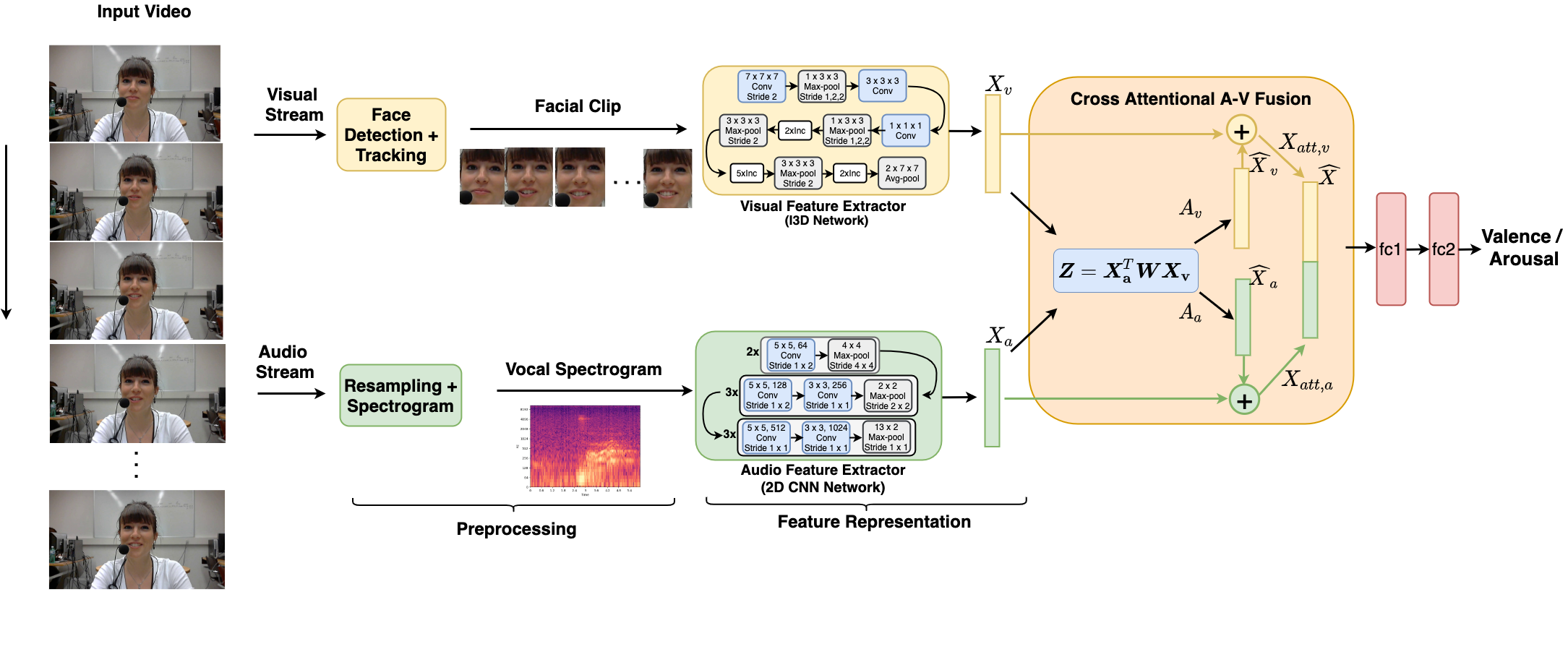}
\caption{\textbf{Block diagram of the proposed cross-attention based A-V fusion model. "Inc" in V feature extractor denotes Inception Module.}}
\label{Block Diagram}
\end{figure*}
\subsection{Visual Network}
Facial expressions from videos involve both appearance and temporal dynamics of the video sequences. Efficient modeling of the spatial and temporal dynamics of the video sequences plays a crucial role in extracting robust features, which in-turn improves the overall system performance. State-of-the-art performance is typically achieved using CNN in combination with Recurrent Neural Networks (RNN) to capture the effective latent appearance representation along with temporal dynamics \cite{7904596}. Several approaches have been explored for dimensional emotion recognition based on LSTMs \cite{5740839}, \cite{WOLLMER2013153}. However, 3D-CNNs are found to be efficient in capturing the spatiotemporal dynamics in videos. Specifically, we consider Inflated 3D-CNN \cite{8099985}, to extract spatiotemporal features of the facial clips from a video sequence.  
Compared to conventional 3D CNNs, I3D can efficiently captures the spatiotemporal dynamics of the V modality while training with less parameters than that of 3D CNNs. Also, it helps to explore the existing pretrained 2D-CNNs, which are trained on many images with facial expressions, thereby improving the spatial discrimination for videos. 
In the proposed approach, we have individually trained the I3D model for the facial modality (see implementation details in Section IV-B). 

\subsection{Audio Network}
The para-lingual information of the speech signal was found to have significant information conveying the emotional state of a person. Even though emotion recognition using voice has been widely explored using the conventional handcrafted features such as MFCC, global features \cite{Sethu2015}, there has been a significant improvement over the recent years with the introduction of DL models. Spectrograms are found to carry significant para-lingual information pertaining to the affective state of a person \cite{Ma2018, Satt2017EfficientER}. Therefore, spectrograms are used in the framework of DL models for speech based emotion recognition. 
Spectrograms have been explored with various 2D CNNs in the literature for emotion recognition \cite{10.1145/3428690.3429153}, \cite{albanie}. We use the A network as shown in Table \ref{speech network} (see  implementation details in Section IV-B). 

\subsection{Cross-Attentional Fusion}
The A and V models have been trained separately and the deep features are extracted for the A and V modalities. The performance of valence and arousal varies significantly for A and V modalities. Due to the rich appearance based information in the V modality, it conveys significant information pertinent to the valence as it depicts the expressions of the sequence. Audio signals carry significant information relevant to the intensity of the expressions, which is efficiently manifested in the energy of A signals. For a given video sequence, the V modality carries relevant information in some video clips, whereas A modality might be more relevant for other clips. Since, multiple modalities convey diverse information for valence and arousal than a single modality, multiple modalities can be effectively leveraged by fusing the A and V modalities in a complimentary fashion. In order to reliably fuse these modalities for the prediction of valence and arousal, we use cross-attention based fusion mechanism to efficiently encode the inter-modal information while preserving the intra-modal characteristics. A block diagram of the proposed model is shown in Figure \ref{Block Diagram}. 

Let ${\boldsymbol X}_{\mathbf a}$ and ${\boldsymbol X}_{\mathbf v}$ represents the deep features of A and V modalities of a given video sequence $\boldsymbol X$, where ${\boldsymbol X}_{\mathbf a}\boldsymbol=\boldsymbol(\boldsymbol x_{\mathbf a}^{ l}\boldsymbol)_{ l\boldsymbol=1}^{L}$ and ${\boldsymbol X}_{\mathbf v}\boldsymbol=\boldsymbol(\boldsymbol x_{\mathbf v}^{ l}\boldsymbol)_{ l\boldsymbol=1}^{L}$. $L$ denotes the number of subsequences of $\boldsymbol X$,  $\boldsymbol x_{\mathbf a}^{ l}$ and $\boldsymbol x_{\mathbf v}^{ l}$ denotes the A and V feature vectors respectively of $l^{th}$ subsequence of the video sequence $\boldsymbol X$. Next, the cross-correlation of A and V features for the subsequences is computed from the given video sequence $\boldsymbol X$ to capture the relevance across the modalities. In order to minimize the heterogeneity between the modalities, a learnable weight matrix $\boldsymbol W\in\mathbb{R}^{K\times K}$ is learned, and the cross-correlation is computed as 
\begin{equation}
\boldsymbol Z={\boldsymbol X}_{\mathbf a}^T\boldsymbol W{\boldsymbol X}_{\mathbf v}
\end{equation}
\noindent where $\boldsymbol Z\in\mathbb{R}^{L\times L}$, $\boldsymbol W$ represents cross-correlation weights among the A and V features and $K$ denotes the feature dimension of the A and V features. 

The cross-correlation matrix $\boldsymbol Z$ gives the measure of correlation among the A and V features. Higher correlation coefficient in matrix $\boldsymbol Z$ shows that the corresponding A and V features of the sub-sequence are strongly related to each other. Therefore, $l^{th}$ column of the cross-correlation matrix $\boldsymbol Z$ shows the correlation measure of $l^{th}$ V feature with $L$ A features. Based on this idea, we compute the cross-attention weights of A and V features, ${\boldsymbol A}_{\boldsymbol a}$ and ${\boldsymbol A}_{\boldsymbol v}$ by applying column-wise softmax of $\boldsymbol Z$ and $\boldsymbol Z^T$, respectively: 
\begin{align}
{\boldsymbol A}_{{\boldsymbol a}_{i,j}}
=\frac{ e^{{\mathbf Z}_{i,j}/T}}{\overset{ K}{\underset{ k\boldsymbol=1}{\sum}} e^{{\mathbf Z}_{k,j}/T}}    
\quad  \text{and} \quad
{\boldsymbol A}_{{\boldsymbol v}_{i,j}}
=\frac{ e^{{\mathbf Z^T}_{i,j}/T}}{\overset{K}{\underset{ k\boldsymbol=1}{\sum}} e^{{\mathbf Z^T}_{i,k}/T}}    
\end{align}
\noindent where $i$ and $j$ represents the $i^{th}$ row and $j^{th}$ column of the cross-correlation matrix $\boldsymbol Z$, and $T$ the softmax temperature.

Since the weights $\boldsymbol W$ are being learned based on the cross-correlation of the A and V features, the attention weights of each modality is guided by the other modality, thereby efficiently leveraging complimentary nature of the A and V modalities.  
After obtaining the cross-attention weights, they are used to obtain the attention maps of the A and V features to make it more comprehensive and discriminative: 
$\mathbf{X}$
\begin{align}
\widehat{\mathbf{X_a}}=\mathbf{X_a}{{\mathbf A}_{\mathbf a}} \quad  \text{and} \quad
\widehat{\mathbf{X_v}}=\mathbf{X_v}{{\mathbf A}_{\mathbf v}}
\end{align}
\noindent where ${{\mathbf A}_{\mathbf a}}$ and ${{\mathbf A}_{\mathbf v}}$ denotes the cross-attention weights of A and V features, respectively. 

The re-weighted attention maps are added to the corresponding features to obtain the attended features:  
\begin{equation}
{\boldsymbol X}_{\mathbf a\mathbf t\mathbf t\boldsymbol,\mathbf a}=\tanh({\boldsymbol X}_{\mathbf a}+\widehat{\mathbf{X_a}})
\end{equation}
\begin{equation}
{\boldsymbol X}_{\mathbf a\mathbf t\mathbf t\boldsymbol,\mathbf v}=\tanh({\boldsymbol X}_{\mathbf v}+\widehat{\mathbf{X_v}})
\end{equation}
The attended V and A features, ${\boldsymbol X}_{\mathbf a\mathbf t\mathbf t\boldsymbol,\mathbf v}$ and $ {\boldsymbol X}_{\mathbf a\mathbf t\mathbf t\boldsymbol,\mathbf a}$ are concatenated to obtain the A-V feature representation, which is given by 
$\mathbf {\widehat X} = [{\boldsymbol X}_{\mathbf a\mathbf t\mathbf t\boldsymbol,\mathbf v} ; {\boldsymbol X}_{\mathbf a\mathbf t\mathbf t\boldsymbol,\mathbf a} ]$.  
Finally, the A-V features are fed  to the fully connected layers for the prediction of valence and arousal.

\section{Experimental Methodology}

\noindent \textbf{A) Dataset:}
The proposed architecture is validated with the REmote COLlaborative and Affective (RECOLA) dataset \cite{6553805}. In total, the data-set consists of 9.5 hours of multimodal recordings, which is recorded by 46 French - speaking participants, performing a collaborative task during a video conference. Among the participants, 17 are French, 3 are German and 3 are Italian. The video sequences are divided into sequences of 5 minutes each, which is annotated with a regressed intensity value for every 40 msec by 6 French speaking annotators (three male and three female). The dataset is split into three partitions: train (16 subjects), validation (15 subjects) and test (15 subjects) by balancing the age and gender of the speakers. Due to the uncontrolled spontaneous nature of expressions of the subjects, the dataset has been widely used by the research community in affective computing for various challenges such as AVEC 2015 \cite{Ringeval:2015}, AVEC 2016 \cite{Valstar:2016}, etc. Most of the existing approaches \cite{cite6}, \cite{cite7} in the literature have validated on the dataset used for AVEC 2016 \cite{Valstar:2016} challenge, which consists of 9 subjects for training and 9 subjects for validation. Therefore, we have also validated the proposed approach on the data-set used in AVEC 2016 challenge. 

\noindent \textbf{B) Implementation Details:}
For the \textbf{V modality}, the faces are extracted and pre-processed from the video sequences of the dataset using MTCNN model \cite{7553523}, which is deep cascaded multi-task framework of face detection and alignment. Faces are resized to $224$x$224$ to be fed to the I3D \cite{8099985} network. In order to generate more samples, the videos of the dataset are converted to sequences of 128 frames with a subsequence length of 16, resulting in 21,284 training samples and 16,177 validation samples. In the proposed approach, inception v-1 architecture is used as the base model, which is inflated from 2D pre-trained model on ImageNet to 3D CNN for videos of facial expressions. For regularizing the network, dropout is used with $p = 0.8$ on the linear layers. The initial learning rate of the network was set to be $1e-4$ and the momentum of $0.9$ is used for Stochastic Gradient Descent (SGD). Also weight decay of $5e-4$ is used. Due to the hardware limitations and memory constraints, the batch size of the network is set to be $8$. Data augmentation is performed on the training data by random cropping, which produces scale invariant model. The number of epochs is set to be 50 and early stopping is used to obtain the best weights of the network.

The \textbf{A network} is composed of 3 blocks of convolutional layers, where the first block has convolutional layer followed by max pooling layer. In the second block, there are two convolutional layers followed by max pooling layer. In third block also, there are two convolutional layers followed by average pooling layer, which gives the feature vectors. Finally, the feature vectors are fed to the linear layers to obtain the final prediction of valence or arousal. All the convolutional and linear layers layers are followed by ReLu activation functions. The speech signal is extracted from the corresponding video sequence and re-sampled to 16KHz, which is further segmented to short speech segments. First, we split the extracted speech signal to 5.12 sec, which corresponds to the sequence of 128 frames of the V network. Next spectrogram is obtained using Discrete Fourier Transform (DFT) of length 1024 for each short speech segment of 5.12 sec, where the window length is considered to be 40 msec and the shift length to be 40 msec in order to match with the granularity of annotation frequency. Following aggregation of short-time spectra, we obtain the spectrogram of 128 x 129. Now a normalization step is performed on the obtained spectrograms. The spectrogram is converted to log-power-spectrum, expressed in dB. Finally, mean and variance normalization is performed on the spectrogram. Apart from mean and variance normalization, no other speech specific processing such as silence removal, noise filtering, etc are performed. These spectrograms are then fed to the deep network depicted in Table \ref{speech network}.
\begin{table}
\begin{center}
      \caption{ Deep Neural Network for Vocal Model. Conv : 64, 5x5, 1x2 denotes a convolutional layer of 64 filters, each of kernel size 5x5 and stride of 1x2. pool : 4x4, 4x4 denotes kernel size of 4x4 and stride of 4x4, Linear : in = 1024, out = 256 denotes linear fully connected layer of input size 1024 and output size 256. All convolutional layers are followed by batch normalization and Rectified Linear Units (ReLu). For linear layer in block 4, we use a sigmoid activation function.}
    \label{speech network}
\begin{tabular}{|c|c|c|c|c|c|} 
	\hline
	 \textbf{Stage}  & \textbf{Layers} & \textbf{Output size}  \\
	\hline \hline
    Input & - &  1 x 128 x 129 \\
	\hline
    \multirow{2}{*}{Block 1}  & Conv : 64, 5x5, 1x2  & \multirow{2}{*}{64 x 31 x 15} \\
      & Max pool : 4x4, 4x4 & \\
	\hline
	    \multirow{3}{*}{Block 2}  & Conv : 128, 5x5, 1x2  & \multirow{3}{*}{256 x 15 x 4} \\
	    & Conv : 256, 3x3, 1x1  & \\
      & Max pool : 2x2, 2x2 & \\
      \hline
      	    \multirow{3}{*}{Block 3}  & Conv : 512, 5x5, 1x1  & \multirow{3}{*}{1024 x 1 x 1} \\
	    & Conv : 1024, 3x3, 1x1  & \\
      & Avg pool : 13x2, 1x1 & \\
      \hline
	Block 4  & Linear : in = 1024, out = 256 & 256x1    \\
	\hline
	Block 5 & Linear : in = 256, out = 1 & 1x1\\
	\hline
\end{tabular}
\end{center}
\end{table}
The A network is trained from scratch, where the initial weights of the network are initialized with values from normal distribution. The number of epochs are set to be 100, and early stopping is used. The network is optimized using SGD with momentum of $0.9$. The initial learning rate is set to be $0.001$ and batch size is fixed to be 16. Due to the limited data, the network might be prone to over-fitting . Therefore, in order to prevent the network from over-fitting, dropout is used with p = 0.5 after the last linear layer. Also weight decay of $5e-4$ is used for all the experiments.  

For the \textbf{A-V fusion network}, we used hyperbolic tangent functions for the activation of cross-attention modules. The dimension of the extracted features of A-V modalities are set to be $1024$. In the cross-attention module, the initial weights of the cross-attention matrix is initialized with Xavier method \cite{pmlr-v9-glorot10a} and the weights are updated using SGD with momentum of 0.9. The initial learning rate is set to be $0.001$ and batch size is fixed to be $16$. Also dropout of $0.5$ is applied on the attended A-V features and weight decay of $5e-4$ is used for all the experiments.

Due to the spontaneity of the expressions of the subjects, the annotations was also found to be highly stochastic in nature. Therefore, a chain of post processing steps are applied to the predictions and labels. A rigorous analysis on some of the post processing steps performed on the annotations was investigated by Huang et al. \cite{Huang:2015}. Tzirakis et al. \cite{cite7} explored a series of post processing steps for validating their architecture on the data-set. Inspired by their approach, we have followed similar series of post processing steps to validate our architecture. (i) median filtering with the window size ranging from $0.4$sec to $20$sec (ii) centering the predicted values by computing the bias between annotated (ground truth) values and predicted values (iii) to match the scaling of predicted values and annotations using the ratio of standard deviation of annotated values and predicted values. (iv) shift in time by shifting the annotations forward in time with values ranging from $0.04$ to $10$sec to compensate for delay in annotations as there can be a delay in correspondence between the annotated values and the video frames. We use a loss function based on Concordance Correlation Coefficient ($\rho_c$) as it has been widely explored for dimensional emotion recognition \cite{cite7}, which is given by: 
\begin{equation}
\begin{array}{l}Loss\;=\;1\;-\;CCC(\rho_c)\end{array}    
\end{equation}

\noindent \textbf{C) Evaluation Metric:}
CCC ($\rho_c$) has been widely used in the literature to measure the agreement level between the predictions ($x$) and ground truth ($y$) annotations for dimensional emotion recognition \cite{cite7}. Let $\mu_x$ and $\mu_y$ represents the mean of predictions and ground truth, respectively. Similarly, if $\sigma_x^2$ and $\sigma_y^2$ denotes the variance of predictions and ground truth, respectively, then $\rho_c$ between the predictions and ground truth is:
\begin{equation}
\rho_c=\frac{2\sigma_{xy}^2}{\sigma_x^2+\sigma_y^2+(\mu_x-\mu_y)^2}
\end{equation}
where $\sigma_{xy}^2$ denotes the predictions -- ground truth covariance.

\section{Results and Discussion}

\noindent \textbf{A) Ablation Study:}
In order to analyze the performance of cross-attention mechanism of the proposed approach, we compare the cross-attention based fusion with various fusion strategies widely used in the literature. One of those fusion strategies is LSTM based fusion, where the A and V features are directly concatenated and fed to the LSTM followed by linear layers. To evaluate LSTM model-based fusion, we have extracted V features from 2D CNN using VGG architecture so that we can have frame level features. The VGG architecture is pretrained on FER data-set similar to \cite{8914655} and further fine-tuned on the RECOLA data-set. Initially, we compare the proposed approach without LSTM, where the A and V features are concatenated and directly fed to linear layers. Then we have used LSTM based fusion by feeding the concatenated features to LSTM layer followed by fully connected layers. Due to the temporal modeling of the concatenated features, the fusion performance improves over the non LSTM based fusion strategy. 

We also compare the performance of I3D with baseline concatenation, where the A-V features are concatenated without attention and fed to linear layers for valence/arousal prediction similar to that of the fusion mechanism of Juan et al. \cite{8914655}. We further compare the proposed approach to that of self-attention \cite{cite8}. Unlike simple feature concatenation, self attention emphasizes on the relevant feature components. Therefore, it improves the system performance compared to simple feature concatenation. However, self-attention does not capture the relevance of correlation across the features of A-V modalities, thereby fails to capture the complimentary nature of the modalities. Since cross-attention based mechanism efficiently captures the complementary nature of the modalities by leveraging the cross-correlation across the A-V features, it was shown that the cross-attention based fusion outperforms most of the widely used fusion strategies.
Table \ref{visual network results} presents the results of our ablation study.  
\begin{table}
    \centering
      \caption{ \textbf{CCC performance of our proposed approach obtained with various components  on the RECOLA dataset. Audio model in Table \ref{speech network} is used to extract A features in all experiments.}}
    \label{visual network results}
\begin{tabular}{|l|c|c|c|c||c|c|c|c|c|c|} 
	\hline
	 \textbf{Method: V + Fusion}  & \textbf{Valence} & \textbf{Arousal} \\
	\hline \hline
    2D CNN + Feature Concatenation                     & 0.538 &  0.680 \\
	\hline
	2D CNN + LSTM                                   & 0.552 & 0.697  \\
	\hline
	I3D + Feature Concatenation                        & 0.579 &  0.732 \\
	\hline
	I3D + Self Attention                            & 0.623 &  0.787 \\
	\hline
	I3D + Cross-Attention (ours)                    & 0.685 &  0.835 \\
	\hline
\end{tabular}
\end{table}

\begin{figure*}[t!]
\centering
\includegraphics[width=0.75\linewidth]{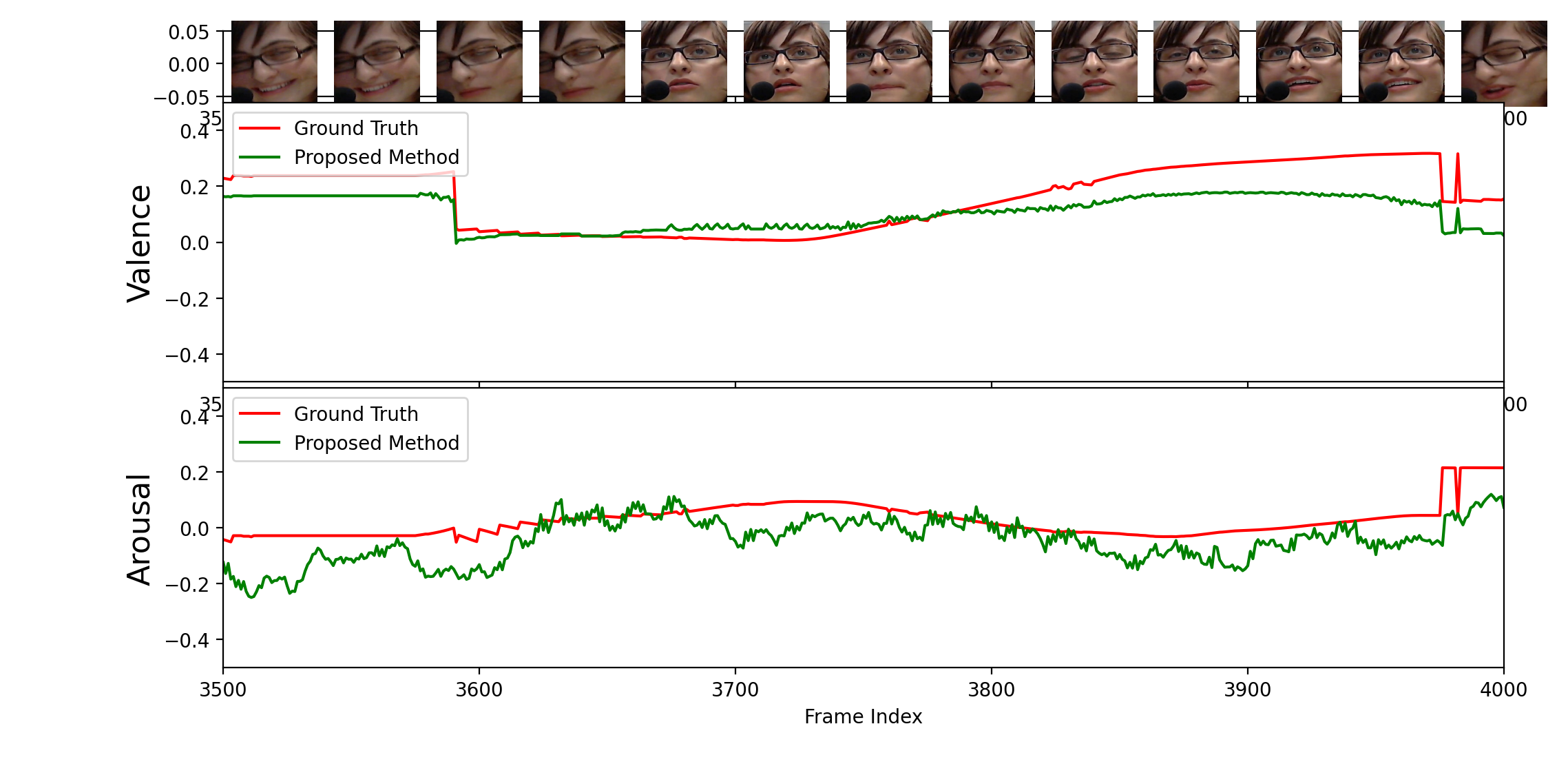}
\includegraphics[width=0.75\linewidth]{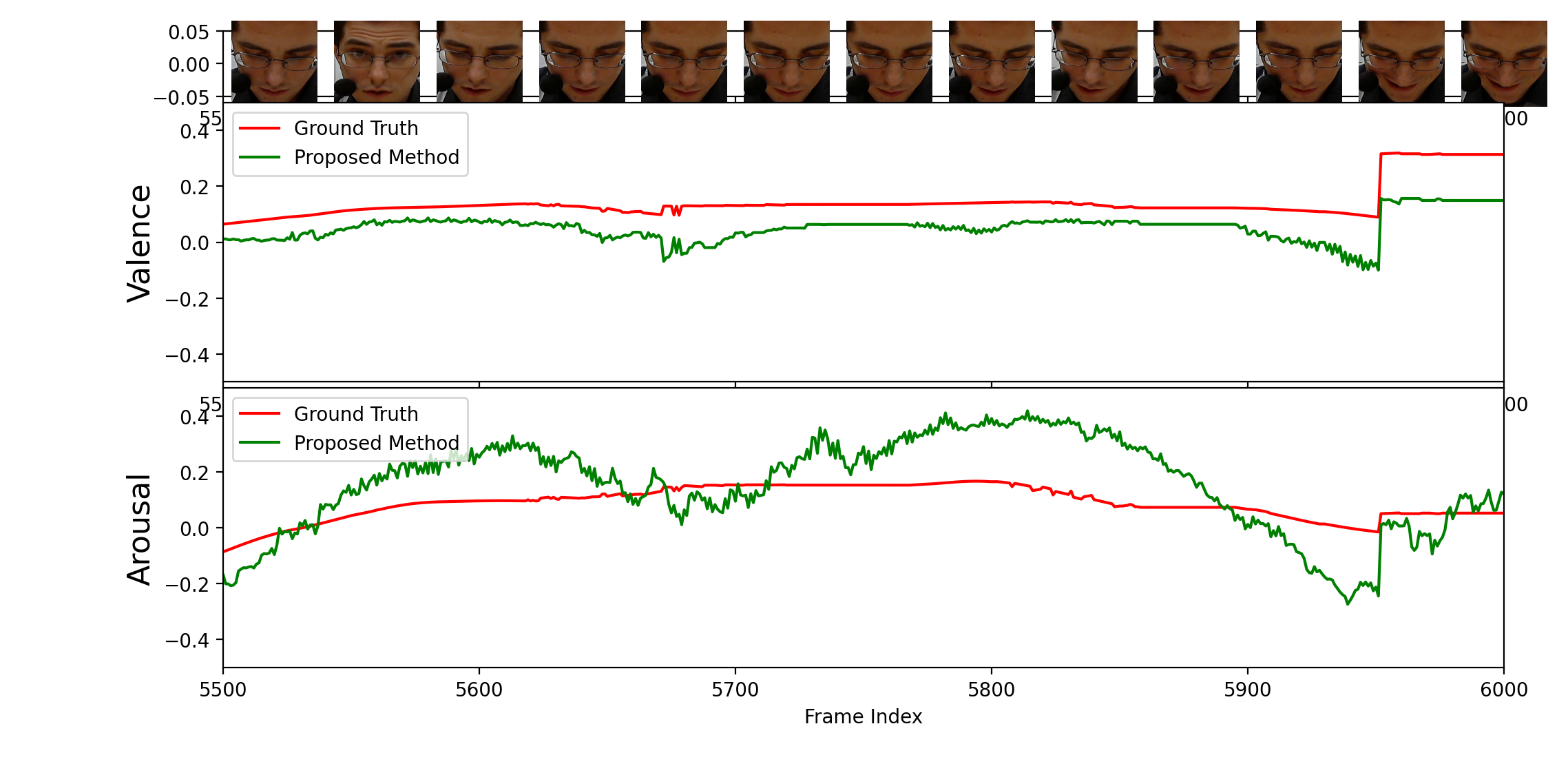}
\caption{\textbf{Visualization of predictions of valence and arousal for subjects "dev 1" and "dev 3" respectively}}
\label{Visual results}
\end{figure*}

\noindent \textbf{B) Visual Results:}
We have further validated the proposed approach by visualizing the predictions of valence and arousal for some of the subjects as shown in Figure \ref{Visual results}. Since the cross-attention mechanism efficiently capture the complimentary nature of the A-V modalities, the proposed approach effectively tracks the ground truth for both valence and arousal. For example, though the full frontal pose of the V modality is not available, the proposed approach still tracks the ground truth of valence and arousal by leveraging the A modality, thereby effectively utilizing the complimentary nature of the A and V modalities. (See Figure \ref{Visual results} lower part)  
\begin{table*}
\renewcommand{\arraystretch}{1.4}
    \centering
      \caption{ \textbf{CCC performance of the proposed and state-of-art models. All the results are presented on the RECOLA development set}}
    \label{Comparison with state-of-the-art}
\begin{tabular}{|l|l||c|c|c|c|c|c|c|c|c|} 
	\hline
	 \multicolumn{2}{|c||}{\textbf{Method}}  &   \multicolumn{3}{|c|}{\textbf{Valence}} & \multicolumn{3}{|c|}{\textbf{Arousal}}  \\ \cline{3-8}
	\multicolumn{2}{|c||}{} & \textbf{Audio}  & \textbf{Visual}  & \textbf{Fusion}  & \textbf{Audio} & \textbf{Visual} & \textbf{Fusion}\\
	 \hline
	\hline
	He et al. \cite{Helang}  & AVEC (2015) & 0.400 &  0.441 & 0.609 & 0.800 & 0.587 & 0.747\\
	\hline
	Han et al. \cite{Han}  & IVU (2017) & 0.480 &  0.592 & 0.554 & 0.760 & 0.350 & 0.685\\
	\hline
	Tzirakis et al. \cite{cite7}  & IEEE JSTSP (2017) & 0.428 &  0.637 & 0.502 & 0.786 & 0.371 & 0.731  \\
	\hline
	Juan et al. \cite{8914655}  & IEEE SMC (2019) & - & -  & 0.565 & - & - & 0.749  \\
	\hline
    Schoneval et al. \cite{cite6} & PR Letters (2021)  & 0.460 &  0.550 & 0.630 & 0.800 & 0.570 & 0.810  \\
	\hline
	Proposed Approach & Cross-Attention & 0.463  & 0.642 & 0.685 & 0.822 & 0.582 & 0.835\\ 
	\hline 
	Proposed Approach & 2-stage Cross-Attention & 0.463  & 0.642 & 0.690 & 0.822 & 0.582 & 0.838\\ 
	\hline
\end{tabular}
\end{table*}
\begin{table}
\renewcommand{\arraystretch}{1.2}
    \centering
      \caption{ \textbf{CCC performance on Fatigue dataset.}}
    \label{Fatigue}
\begin{tabular}{|l||c|c|c|c|c|c|c|c|c|c|} 
	\hline
	 \textbf{Method}  
	 & \textbf{Fatigue Level} \\
	 \hline
	 Audio only (2D-CNN) & 0.312 \\
	\hline
	Visual only (I3D) & 0.415\\
	\hline
	Feature Concatenation & 0.378 \\
	\hline
    Proposed Approach (Cross-Attention) & 0.421 \\
	\hline
\end{tabular}
\end{table}

\noindent \textbf{C) Comparison with State-of-the-Art:}
Conventional ML approaches have been explored based on hand-crafted features for dimensional emotion recognition. He et al. \cite{Helang} explored handcrafted features (LPQ-TOP) for V and Low level descriptors (LLD) such as MFCC, energy, etc. for A, along with physiological modalities including electro-cardiogram (ECG) and electro-dermal activity (EDA).
Due to the use of additional physiological modalities and more LLD descriptors in A, as well as additional geometric features of V, the fusion performance has significant improvement. Han et al. \cite{Han} explored LLD features for A and facial landmark features (only geometric) for V, and fused in a hierarchical fashion by leveraging the individual advantages of SVM and neural networks and showed further improvement in valence. 
Inspired by performance of DL models, Tzirakis et al. \cite{cite7} explored deep models in an end-to-end fashion by using Resnet50 for V and 1DCNN for A on raw data. However, the features are directly concatenated and fed to 2 LSTMs, thereby resulting in a decline of fusion performance over individual modalities. The performance has been further improved by Juan et al \cite{8914655}, where they have used pretrained model on FER for V and LLD for A. In most of the above mentioned approaches, A-V features are directly concatenated and fed to the prediction model. However, it was found that direct concatenation of features fails to capture the inter-modal relationships \cite{8269806}. Recently, Schoneval et al. \cite{cite6} used knowledge distillation for V and VGG network using spectrograms for A. Instead of direct concatenation, they have used two independent CNNs before concatenating them and showed that their fusion performance has improved over the performance of individual modalities. Though deep models have improved the performance over handcrafted features, they fail to effectively leverage the complimentary nature of the A-V modalities. By leveraging the cross-correlation of A and V features, we have improved the system performance using cross-attention based fusion. Therefore, the proposed approach was found to outperform state-of-the-art approaches. For the sake of completeness, we have also compared the proposed approach with 2-stage cross-attention as in \cite{lee2021crossattentional}. Though it shows slight improvement in performance, the proposed approach with single iteration outperforms state-of-the-art for dimensional emotion recognition with less computational complexity compared to 2-stage cross-attention. 

\noindent \textbf{D) Results with Fatigue Dataset:}
The proposed approach was also validated on Fatigue (private) dataset. The Fatigue dataset is obtained from $18$ participants in Rehabilitation center, who are suffering from degenerative diseases inducing fatigue which affect their life quality. A total of 27 video sessions are captured from $18$ participants with a duration of 40 - 45 minutes and the videos are labeled at sequence level on a scale of 0 to 10 for every 10 to 15 minutes. We have considered $80\%$ of data as training data (50,845 samples) and $20\%$ as validation data (21,792 samples). The results are presented on the validation data. Due to the rich information of arousal in A, the performance of arousal performs better than valence in A. Similarly, the performance of valence is better than arousal for V modality. We have also compared our proposed approach with the baseline feature concatenation without cross-attention. Finally, the performance of the proposed approach with cross-attention module is presented. It was found that the proposed approach significantly improves the system performance over baseline of feature concatenation as shown in Table \ref{Fatigue}.

\section{Conclusion}
In this paper, we introduce cross-attentional A-V fusion for  dimensional emotion recognition from videos. Unlike prior approaches, we focus on improving the fusion strategy based on cross-correlation across the A-V features. First, DL models are trained individually for V and A modalities, where features are extracted using and I3D and 2D-CNN, respectively. Then, an attention mechanism based on cross-correlation between A and V features are applied on the individual modalities. Finally, the attention weighted features are concatenated and fed to two linear connected layers for predicting valence and arousal. The proposed model efficiently combines the modalities in a complimentary fashion, and significantly improves the performance of the system. Experiments show that our cross-attentional A-V fusion provides competitive results w.r.t. to state-of-the-art approaches.


{\small
\bibliographystyle{ieee}
\bibliography{CameraReadyVersion}
}

\end{document}